# A Review of Full-Sized Autonomous Racing Vehicle Sensor Architecture

Manuel Mar[1], Vishnu Pandi Chellapandi[2], Liangqi Yuan[2], Ziran Wang[2], and Eric Dietz[1]

*Abstract—* In the landscape of technological innovation, autonomous racing is a dynamic and challenging domain that not only pushes the limits of technology, but also plays a crucial role in advancing and fostering a greater acceptance of autonomous systems. This paper thoroughly explores challenges and advances in autonomous racing vehicle design and performance, focusing on Roborace and the Indy Autonomous Challenge (IAC). This review provides a detailed analysis of sensor setups, architectural nuances, and test metrics on these cutting-edge platforms. In Roborace, the evolution from Devbot 1.0 to Robocar and Devbot 2.0 is detailed, revealing insights into sensor configurations and performance outcomes. The examination extends to the IAC, which is dedicated to high-speed self-driving vehicles, emphasizing developmental trajectories and sensor adaptations. By reviewing these platforms, the analysis provides valuable insight into autonomous driving racing, contributing to a broader understanding of sensor architectures and the challenges faced. This review supports future advances in full-scale autonomous racing technology.

## I. INTRODUCTION

The Society of Automotive Engineers (SAE) has classified autonomous driving into six levels, ranging from basic driver assistance to full automation. At level 0, there is no automation, while level 5 represents complete autonomy, eliminating the need for human intervention [1]. These systems are highly dependent on a variety of sensors and software to accurately perceive their surroundings, to achieve full automation without human intervention. As we move closer to this reality, the sensor industry is experiencing rapid growth and innovation to meet the challenges that autonomous systems present [2]–[4]. In particular, autonomous ground systems such as cars, trucks, and trains continue to grapple with specific unresolved challenges, fueling the motivation of researchers and engineers to dive into this area [5], [6]. The world of motorsports, characterized by conditions such as steep inclines, high-speed corners, and nuanced techniques such as "lift and coast", presents its unique set of challenges. Racing circuits featuring vehicles such as IndyCar, Formula E, and Formula 1 represent the pinnacle of high-performance design [7]. Technological innovations nurtured in these racing arenas often find their way into commercial vehicles [8]. Racing drivers, with their deep understanding of vehicle dynamics and performance, exhibit skills and techniques that are difficult to replicate through software or automated systems. Although sensors can process information faster than human senses, the nuanced understanding a racer possesses often exceeds that of an average driver. In recent times, numerous autonomous racing competitions have emerged that challenge researchers [9]–[13].

High-speed racing vehicles, as defined within the context of this research, encompass vehicles engineered to operate in stringent conditions, experience substantial lateral forces, and are capable of achieving rapid acceleration. Technological advances in sensors have led to a variety of high-performance ground vehicle programs, each characterized by different metrics, propulsion systems, and performance results [14]. The last decade has introduced multiple autonomous racing programs in which ground race vehicles were built specifically for racing purposes, where vehicles are handled at their limits [15]. Various competitions have been held, but this research will concentrate on two significant projects that involved the development of full-sized autonomous racing vehicles for racing platforms: Roborace and the Indy Autonomous Challenge (IAC). Although sharing similar conceptual foundations, both programs diverge in aspects like race architecture, sensor suite, main propulsion source, and performance metrics.

Autonomous hardware for this research will be referred to as components in the vehicle that contribute to vehicle autonomy mobility, such as exteroceptive sensors (e.g., LiDAR, radars, cameras), localization (GNSS, GPS, IMU), main computer system onboard, and the network system (e.g., V2I, V2V, C-V2X) [16]–[19]. Previous studies [20]–[24] have explored the autonomous racing domain, delving into aspects such as vehicle dynamics, simulation, and lessons learned from operating racing vehicles. Building on this foundation, our review expands into an examination of hardware and software limitations observed during recent full-scale races. This comprehensive overview highlights research and efforts in the realm of full-size autonomous racing vehicles, which present significant contributions to the field.

1) We provide a comprehensive display and overview of the current autonomous hardware stack of full-size autonomous vehicles, detailing sensor characteristics and overall architecture used in each race.
2) We conduct a systematic review of research papers documenting hardware and software malfunctions during events, categorizing the challenges and problems encountered with each sensor.
3) We present an analysis of existing challenges for sensors and future research directions in the context of racing cars, encompassing real-time responsiveness, standardization, sensor fusion, and deep learning modeling.

[1]M. Mar and E. Dietz are with the Polytechnic Institute, Purdue University, West Lafayette, IN 47907, USA. Emails:{mmar, jedietz}@purdue.edu
[2]V. P. Chellapandi, L. Yuan, and Z. Wang are with the College of Engineering, Purdue University, West Lafayette, IN 47907, USA. Emails: {cvp, liangqiy, ziran}@purdue.edu

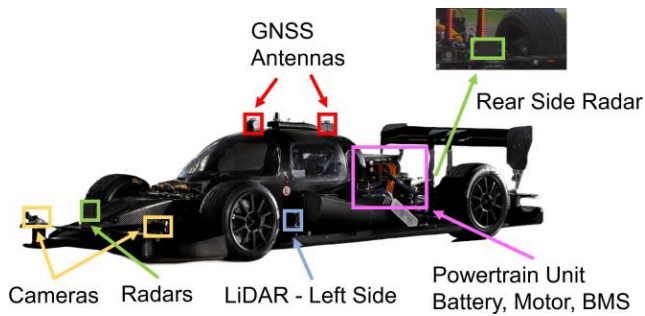

Fig. 1. Devbot 1.0 Sensor Architecture

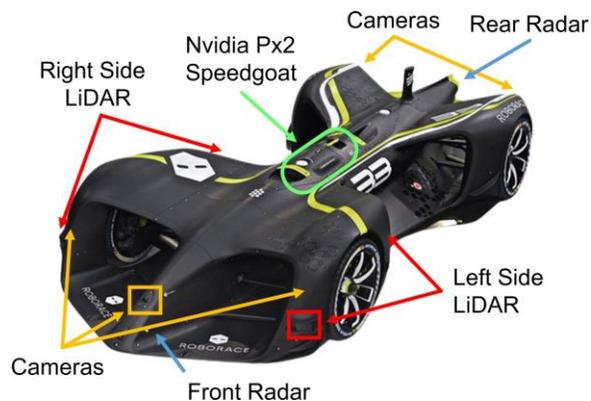

Fig. 2. Robocar Sensor Architecture

This study primarily focuses on examining the vehicle architecture design, evaluating the sensor suite, and presenting an overview of the autonomy hardware architecture's performance.

## II. ROBORACE

Roborace was an autonomous vehicle racing series founded in 2016 to showcase the capabilities of autonomous systems technology and artificial intelligence in an electric-powered vehicle [9]. There have been three different built versions for this race series: Devbot (2016), Robocar (2017), and Devbot 2.0 (2018). In this section, we describe the vehicle architecture along with some of the testing metrics that were obtained from this project.

### A. Sensor Setup

Roborace vehicles have more similarities than differences as shown in Table II, the main difference between each version is the presence of a cockpit, aerodynamic design, and sensor placement.

### B. Devbot 1.0

Roborace unveiled its inaugural autonomous vehicle, named Devbot. Constructed around a Le Mans Prototype (LMP) chassis, with four 135kW electric power each wheel for a combining 500 HP, without an engine cover for better cooling and access [9]. The vehicle incorporated a cockpit seat in case human intervention tests were needed, this vehicle was intended to test all the hardware functionality before launching the latter vehicle racing series such as Robocar and Devbot 2.0. The sensor suite incorporated into Devbot comprised computational power onboard among two primary electronic control units (ECUs). The NVIDIA Drive PX2 was responsible for high-level planning and decision-making, while the Speedgoat Mobile Target Machine catered to the real-time control tasks [53]. This vehicle also incorporated LiDARs, cameras, radars, and ultrasonic speed sensors that placement and make are shown in Figure 1 and Table II.

### C. Robocar

The Robocar was built with a different concept design, it does not have a cockpit, therefore, there was no manual override for a racer or driver, the intention of this update was to run at even higher speeds compared to its predecessor by removing the weight of a driver. The competition format with this vehicle consisted of giving a university team, a fully autonomous vehicle to run their software stack, this vehicle had an identical sensor suite as Devbot 1.0, two front and four surround cameras, five LiDARs, two radars in the front and back, and 17 ultrasonic sensors [53], the Robocar sensor architecture is presented in Fig. 1.

The control system's design is bifurcated into two categories: non-real-time architectures featuring GPUs and real-time architectures. The latter integrates the Speedgoat operating at 250 Hz, the IMU, and the differential GPS also at 250 Hz, while the LiDARs operated at 25 Hz [54], [55]; it also had a McLaren ECU on it [56]. Some results from one of the first tests done with this vehicle are listed below.

- **Computing Issues**: The scan matching in PX2 overloads the ARM CPU, leading to occasional failures [54].
- Vehicle Alignment Concerns: The single mass model used for trajectory planning does not account for the vehicle's alignment not always being tangent to the path, leading to lower acceleration in some turns [54].
- **Localization Error**: Below 0.3 meters, though improvements are possible with better algorithms or techniques [54].
- **Safety Limitations**: A maximum speed limit of 50 km/h and maximum normalized longitudinal and lateral accelerations of 0.8 [54].

### D. Devbot 2.0

The Devbot 2.0, unveiled in late 2018, was not an improvement of the Robocar, but rather an enhancement of Devbot 1.0. Its sensor configuration mirrored its predecessors, using both Speedgoat and NVIDIA PX2 as core components. Higher speeds were reached with this vehicle, where previous experience in racing and crashing provides

TABLE I
Major Milestones during Races and Malfunctions

| Race | Reference | Year | Vehicle | Sensor | Method | Task and Outcome |
|---|---|---|---|---|---|---|
| Roborace | Nobis et al. [25] | 2017 | Devbot 1.0 | LiDAR | SLAM | **Localization and Mapping.** Accurately determining the positions and orientations of four LiDAR sensors, leading to small angular offsets in sensor readings during data fusion. |
| | Stahl et al. [26] | 2019 | | LiDAR | Kalman Filter | **Localization.** Struggles with longitudinal estimation at high speeds due to the absence of unique features on straight race tracks and time delays in LiDAR data processing. |
| | Heilmeier et al. [27] | 2019 | | LiDAR + IMU + GPS | Quadratic Optimization | **Trajectory Planning.** Robust implementation of planning and control at 150km/h and 80 % acceleration limits. Slow for online implementation. |
| | Wischnewski et al. [28] | 2019 | | LiDAR + GPS | Kalman Filter | **Localization and State Estimation.** Estimation residuals are zero mean, giving a high-performance controller at top speeds of 150 km/h. |
| | Wischnewski et al. [29] | 2019 | Devbot 2.0 | Vehicle Status | Gaussian Process | **Trajectory Planning.** Vehicle cannot adjust its acceleration limitation to different circumstances in different areas on the track. |
| | Betz et al. [30] | 2019 | | Vehicle Status | TUM Roborace | **Crash Explaination.** Velocity Planner Failures; Controller failed to switch to the emergency trajectory; Behavior planner communication 300 ms timeout; Maximum lateral error of 5 meters. |
| | Stahl et al. [31] | 2019 | | Vehicle Status | Graph Learning | **Trajectory Planning.** Online planner tested in a physical vehicle with success at 212 km/h |
| | Hermansdorfer et al. [32] | 2020 | | Vehicle Status | TUM Roborace + KPI | **Path Planning.** Software is more conservative for velocity, acceleration limits and safety checks, while a driver reaches the handling limits by applying more limits acceleration / deceleration rates. |
| | Renzler et al. [33] | 2020 | | LiDAR | Distortion Correction | **Localization.** Successfully implemented in-vehicle platform, issue with LiDAR readings is solved. |
| | Massa et al. [34] | 2020 | | LiDAR + IMU + GPS | Kalman Filter | **Localization and State Estimation.** Pointcloud noise, LiDAR calibration error, algorithm parameters tuning. |
| | Herrmann et al. [35] | 2020 | | Vehicle Status | Sequential Quadratic Programming | **Vehicle Control.** Tested Velocity optimization algorithm up to 200 km/h in hardware-in-loop simulation. |
| | Hermansdorfer et al. [36] | 2021 | | Vehicle Status | Gated Recurrent Unit Network | **Dynamics Modeling.** Model fails to extrapolate driving situations, more date required |
| | Schratter et al. [37] | 2021 | | LiDAR + GPS | Normal Distributions Transform | **Localization and Mapping.** The used method for localization overloads the used CPU reaching 100%. |
| | Christ et al. [38] | 2021 | | Vehicle Status | Vehicle Dynamic Model | **Trajectory Planning.** Tested on simulation and physical vehicle tests. |
| | Herrmann et al. [39] | 2022 | | Vehicle Status | Sequential Quadratic Programming | **Vehicle Control.** Tested successfully in CPU (Simulation), solved the problem in less than 15 seconds with real-time data. |
| IAC | Lee et al. [40] | 2022 | AV-21 | GPS + LiDAR | Kalmant Filter | **Path Planning.** Successfully implemented, nonetheless, GPS presented critical degradation during the test which is warned safely, and the wall following navigation was activated. |
| | Wischnewski et al. [41] | 2022 | | Vehicle Status | Model Predictive Control (MPC) | **Vehicle Control.** Top speed achieved 265 km/h and lateral accelerations up to 21m/s$^2$ on the Las Vegas Motor Speedway (LVMS). |
| | Meyer et al. [42] | 2022 | | LiDAR | Clustering | **Wall Curvature Detection.** Wall detection worked even in the presence of occlusions, successfully deployed in an oval track. |
| | Schmid et al. [43] | 2022 | | Vehicle Status | MPC | **Vehicle Control.** Description of vehicle architecture with some of the faced constraints. |
| | Karle et al. [44] | 2022 | | Vehicle Status | LSTM-based Encoder-Decoder | **Trajectory Planning.** Tested in a Hardware-in-the-Loop (HIL) simulator and vehicle platform. |
| | Seong et al. [45] | 2022 | | Vehicle Status | MI-HPO Model | **Vehicle Control.** Tested on track at Indianapolis Motor Speedway (IMS) and LVMS at speeds over 200 km/h. Model identification showed optimized dynamic parameters offline. |
| | Raji et al. [46] | 2022 | | LiDAR+GPS | Nonlinear MPC | **Vehicle Control.** Tested on track at top speeds of 167 mph. The controller tried to reach a higher speed but could not due to a powertrain issue. |
| | Spisak et al. [47] | 2022 | | GPS | Linear Quadratic Regulator (LQR) | **Vehicle Control.** Tested at top velocities of 65 m/s in an oval track with low error ($< 0.5$ m), stability is guaranteed at 140 mph. Empirical tuning of the controller is not optimal. |
| | Betz et al. [48] | 2023 | | Camera + LiDAR + Radar + GPS | Autonomous System | **Full Self-Driving.** Critical variable end-to-end latency must be low and stable for developing robust autonomous driving software. |
| | Jung et al. [21] | 2023 | | Camera + LiDAR + Radar + GPS | Autonomous System | **Full Self-Driving.** Speeds up to 220 km/h were reached in a solo lap, and an acceleration pf 12.41 m/s2 was reached. Lack of generalization due to the closed environment and rules of the competition. |
| | Trauth et al. [49] | 2023 | | Vehicle Status | TUM Autonomous Motorsport | **Full Self-Driving.** Deployed in IAC race, a gradient-free algorithm based on testing optimization leads to efficient use of real testing time by ensuring desired software configuration. |
| | Raji et al. [50] | 2023 | | Camera + LiDAR + Radar + GPS | Autonomous System | **Full Self-Driving.** The head-to-head race crash resulted from a safety check setting the threshold for heading error too strictly at 6 degrees, not accounting for extreme real-world scenarios. |
| | Raji et al. [51] | 2023 | AV-21-Refresh | Vehicle Status | MPC | **Vehicle Control.** The proposed model presents less error than the classic single-track model within the scenario of a race course track, with limited testing time for tuning. |
| | Lee et al. [52] | 2023 | | Camera + LiDAR + Radar + GPS | Kalman Filter | **Localization and State Estimation.** Tested in a vehicle. Even if there is degradation or loss of GPS/INS or an undesired given path, the proposed system reliably counteracts safely. |

TABLE II
ROBORACE VEHICLES AUTONOMY ARCHITECTURE

| Series | 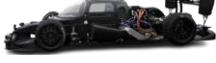 | 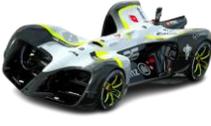 | 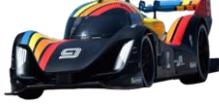 |
|---|---|---|---|
| | **Devbot 1.0** | **Robocar** | **Devbot 2.0** |
| **Release Year** | 2016 | 2017 | 2018 |
| **Real Time Processor** | Speed Goat Target Machine | | |
| **Non real time Processor** | Nvidia Drive PX2 | | |
| **Chassis** | LMP3 | Daniel Simon | LMP3 |
| **LiDAR** | 5 × Ibeo ScaLa B2 s | 5 × Ouster OS-1-64/16 | 5 × Ouster OS-1-64/16 |
| **GPS/IMU** | OxTS RT 4000 | OXTS RT 2 | OxTS RT 4000 |
| **Camera** | 6 × - | | |
| **Radar** | 2 × - | | |
| **Optical Speed Sensor** | Kistler SFII | - | Kistler Correvit SFII |
| **Gyroscope** | - | - | Mclaren – 250Hz |
| **Top Speed** | 186 km/h | 282.60 km/h | 21 km/h2 |

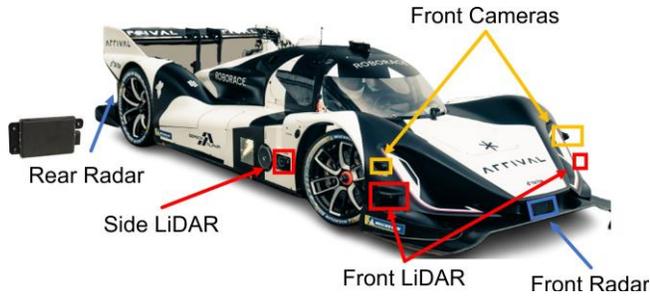

Fig. 3. Devbot 2.0 Sensor Architecture

more insight for later software and firmware updates on this generation. The tests were held in different circuits: Zala-Zone, Hungary; Circuit de Croix-en-Ternois, France; Berlin Formula E racetrack, Germany and Monteblanco, Spain.

Most of the teams used Ouster OS1 LiDAR, capable of 360-degree scans and 1024-point resolution for localization. The mapping took advantage of 64-layer OS1-64, while 16-layer OS1-16 helped with localization. All sensors were positioned at a height of 1.1m, including two GPS antennas. They also used the Oxford RT 4000 as a reference measurement system. The TUM team vehicle planner operated at an average rate of 16.8 Hz, and its capabilities were evaluated up to 212 km/h with a prediction of 200 ms to anticipate the movements of the leading vehicles [31].

During races, no GPS data under 20 cm was accessible, and the LiDAR-based localization's lateral error remained under 10 cm. The vehicle achieved speeds greater than 45 m/s and accelerations greater than 10 m/s². Time synchronization with the sensors was a challenge. For example, at 30 m/s, a 10 ms delay could lead to a 0.3 m error, which requires vehicle odometry [37]. In Zala, the Pisa Team made strides in the correction of the distortion of lithium and the compensation of the delay, attaining speeds of up to 90 km/h and lateral and longitudinal accelerations of 10 m/s² at a rate of 20 Hz [33]. Furthermore, the teams tested Kalman filters in conjunction with LiDAR, IMU, and vehicle dynamic sensors. They reached a maximum speed of 90 km/h. However, due to noise and delays in IMU acceleration measurements, they were not considered for estimation. During Roborace's Alpha Season, Graz University took the lead and clocked the fastest lap time of 1 minute and 37.440 seconds at the Circuit de Croix-en-Ternois, averaging around 65 km/h [57].

A few of the performance and metrics parameters for this vehicle were:

- **State Estimation**: The enhanced H-infinity filter, which relied on-vehicle sensors such as LiDAR, IMU, GPS, and Vehicle Odometry, maintained consistent estimation errors across laps, compared to EKF showed increased errors after each lap [57].
- **Implications of High-Speed LiDAR Use**: The experiment showed that even at high speeds, with longitudinal and lateral accelerations of up to 10 m/s², accurate LiDAR measurement and correction can be achieved [33].
- **Effect of Distortion on Dynamic Driving**: The distortion was less visible because the reflections were within the boundaries of the track. The difference between distorted and corrected point clouds gradually decreased from the first to the fourth quadrant [33].
- **Vehicle Alignment Concerns**: The single mass model used for trajectory planning does not account for the vehicle's alignment not always being tangent to the path, leading to lower acceleration in some turns.
- **Localization Error**: Below 0.3 meters, though improvements are possible with better algorithms or techniques.
- **Acceleration Discrepancies**: Differences observed in high acceleration values, possibly due to conservative estimations in the handling map or unaccounted for suspension dynamics.

TABLE III
INDY AUTONOMOUS CHALLENGE (IAC) VEHICLE COMPARISON DIFFERENCES

| Series | AV-21 | Dallara AV-21 Refresh |
|---|---|---|
| Release Year | 2021 | 2022 |
| Computer | Adlink – AVA 3501 | Autera-Autobox |
| GPU | NVIDIA Quadro RTX 8000 | NVIDIA RTX A5000 |
| Chassis | Dallara | Dallara |
| LiDAR | Luminar Hydra | Luminar Hydra |
| GPS/IMU | Novatel Powerpak | Novatel Powerpak - Vectornav |
| Camera | Mako | Mako |
| Radar | Aptiv (ESR-MRR) | Aptiv (ESR-MRR) |
| Top Speed | 281 km/h | 276 km/h |

## III. INDY AUTONOMOUS CHALLENGE

The Indy Autonomous Challenges was founded to provide university teams with the opportunity to race high-speed self-driving vehicles against each other. This vehicle was initially developed and prototyped through the Deep Orange 12 project at Clemson University using the Dallara IL-15 chassis [58]. There have been two major versions of IAC, the first initial deployment which was released in 2021 (Dallara AV-21), and the second version which was released after the second semester of 2022. The general sensor architecture is shown in Fig. 3. Similar to Roborace, updates were required to adapt to the demands of high-speed racing. The primary modifications to the autonomous sensor stack improved the reliability of the localization system and the incorporation of an additional GNSS unit to improve system redundancy. Additionally, considering the substantial data flow from various sensors, a revised network configuration was imperative to alleviate data congestion within the vehicle.

### A. Sensor Challenges

*1) LiDAR:* LiDAR faces challenges, including its high cost, limitations in mechanical scanning, susceptibility to disturbances from external light sources, and safety restrictions for the human eye, which limit its detection distance to approximately 100 meters [59]. The LiDAR used for this project was from Luminar, the Hydra model [60]. Various challenges associated with LiDAR were identified:

- **Delay due to Reflection**: A high number of reflections induced a delay in the LiDAR perception process, leading to complications in object recognition, particularly at high speeds [61].
- **Banking Angle Adaptation**: The hardware was modified to maintain a narrow opening angle on straight paths and to expand its field of view (FOV) during turns. This adjustment was essential due to the constraints of its restricted vertical FOV, especially with fluctuating banking angles.

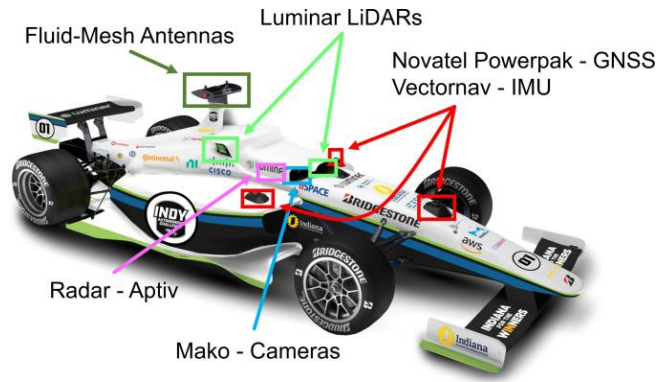

Fig. 4. AV-21 Sensor Architecture

- **Scan Matching Issue**: A LiDAR-based global position method cannot find a solution to match the scan while driving at high speed; therefore, two GPS measurements had to be integrated [62].
- **Driver Crash**: The LiDAR driver crashed during the start-up process which caused the LiDAR to report the last before [63].
- **Point-cloud Density**: LiDAR provides a great tool with 3D images; however, point clouds need to be processed and filtered to reduce the number of points; computation time for TUM was claimed to be 22ms, and preprocessing achieved less than 20% of issues [61].

Some performance and results have shown promising results of the robustness of the sensors despite the high vibrations and the conditions of the scenario. For example, in [42], a high-precision wall detection with less than 0.11 m of error; they could handle occlusions and provide consistent wall detection, even with interferences with a process time analysis of 15 Hz.

*2) IMU and GPS:* Localization has historically presented challenges, but over the years, various solutions have been developed and implemented [64]. In the context of highly dynamic scenarios involving ground vehicles, there has been a paucity of academic research that addresses the resilience of these systems under substantial lateral forces. Specifically, within the domain of autonomous ground vehicle racing, numerous teams relied on conventional localization methods, including extended Kalman filters (EKF) and Sensor Fusion, as well as preexisting packages like Autoware's Robot Localization, which integrates wheel odometry and IMU data. Some attempts have also been made to incorporate LiDAR as a backup localization source. However, due to computational demands, the reliability of LiDAR at speeds exceeding 100 mph remains a concern. To succinctly summarize, the following sections detail the challenges encountered and the successes achieved on the racetrack.

- **Data Filtering**: Different edges can cause inaccurate GPS data, which caused early crashes during testing, for example, a team crashed because of the GPS output data showing that the vehicle rotated 90 degrees between two data points [65].
- **IMU-GPS Continuous Acceleration**: In the downward direction (z-axis), incorrect data led to a continuous increase in velocity estimation within the GPS system. This eventually triggered export controls, halting all GPS output [65].
- **Update Failure**: A spin was caused because the GPS did not provide an update to the state estimator, causing a cumulative integration of errors [61].
- **Severe Vibrations**: There was multiple positioning degradation of the two GNSS units of the first AV-21 version due to strong vibration [40].
- **Need of Cellular or Internet Connectivity**: The lack of cellular connectivity introduced several issues such as the RTK would not receive the correction values in some areas [65], which is a problem in the remote testing track or specific areas of a large track.

3) *Powertrain:*

- **Hardware Limitations**: During Indianapolis Motor Speedway (IMS) and Las Vegas Motor Speedway (LVMS) race events, vehicle speed was restricted due to hardware limitations and engine malfunctions, preventing it from achieving the desired target speed even when the throttle was fully engaged [46].
- **ECU Latency**: A wrong hard brake command was triggered by a hardware ECU module that is not related to the motion planner and controller [46].
- **Tuning Issues**: Speed was limited due to a cable that was attached to the powertrain system being disconnected due to setting the limp mode of the engine. Additionally, the controller requested full throttle during race time and there was an oscillation of throttle due to a non-ideal tuning of the turbocharges and a malfunction of its mechanic [66].

4) *Radar:*

- **Radar Noise**: The radar has more sensor noise than the LiDAR, where the radar sees higher deviations [67].
- **Filter Adjustments**: Adjustment and fine-tuning of the radar for prone areas. Measurement of object velocity and its high sensor range was fundamental for overtaking scenarios [61].

## IV. OPPORTUNITIES AND FUTURE DIRECTIONS

### A. Adaptive Algorithms for Real-Time Performance

Develop and implement adaptive learning algorithms that enable real-time adjustments to sensor configurations and the overall autonomy system. By allowing vehicles to continuously learn and adapt to the intricacies of different racing scenarios, these algorithms could significantly improve performance and responsiveness. The focus should be on creating dynamic systems that leverage machine learning to optimize decision making based on real-time data, ultimately improving racing efficiency and competitiveness [68], [69].

### B. Standardization in Autonomous Racing Hardware

Future research could advocate for standardization of sensor interfaces, communication protocols, and data formats on different autonomous racing platforms. Establishing industry-wide standards would facilitate interoperability, encourage knowledge sharing, and accelerate the development of robust and versatile sensor architectures. This initiative could lead to a more collaborative ecosystem, where advancements in one platform can benefit the entire autonomous racing community, fostering innovation and accelerating progress in the field [70], [71].

### C. Sensor Fusion Optimization

Integration of multi-modal sensors – LiDAR, cameras, radar, GPS, and IMU are being actively researched to enhance perception reliability, aiming for a comprehensive and precise representation of the vehicle's surroundings [72]. As the field evolves, future opportunities lie in developing dynamic sensor calibration methods to adapt to changing sensor positions, ongoing research in real-time processing techniques to achieve low-latency sensor fusion in dynamic environments, and the implementation of redundancy strategies to improve system robustness [73]–[76]. Additionally, adaptive filtering approaches, such as Kalman filters and particle filters, offer opportunities to dynamically adjust parameters, ensuring optimal sensor fusion performance under various conditions.

### D. Complex Dynamic Modeling through Deep Learning

Deep learning is widely used to model complex dynamics by processing sensor data sequences due to their ability to improve generalization in diverse driving scenarios, ensuring adaptability and reliable performance [77], [78]. Leveraging transfer learning, particularly in autonomous racing scenarios, presents an opportunity to efficiently extrapolate knowledge from pre-trained models to specific conditions. The focus on data enhancement strategies is crucial to increase the diversity of training data sets, simulate environmental variations, and contribute to the development of more robust models. Furthermore, the exploration of uncertainty estimation techniques, such as Bayesian approaches or ensemble methods, holds promise for enhancing the reliability of prediction modeling.

## V. CONCLUSIONS

This review paper examines and highlights key milestones and challenges encountered in the development of full-size autonomous racing vehicles and their sensor architectures. Through a comprehensive analysis of past and ongoing research endeavors, we have identified the algorithms, hardware components, and overall full-stack architectures that have been specifically implemented for full-size autonomous competitions and subsequently tested on racetracks. Our presentation delves into the categorization of this research, shedding light on the challenges, objectives, and significant outcomes obtained from these investigations. We are specifically focusing on the implementation of distinct software stacks in real-time industrial ground autonomous vehicles. Looking ahead, future studies will explore various-sized autonomous race vehicles, providing benchmarks and conducting more thorough comparisons of hardware performance.